\documentclass[10pt,twocolumn,letterpaper]{article}

\usepackage[pagenumbers]{iccv} % To force page numbers, e.g. for an arXiv version
\usepackage{multicol}
\usepackage{multirow}
\usepackage{makecell}
\usepackage{pifont} % xmark
\usepackage{colortbl} % rowcolor
\usepackage[T1]{fontenc}
\usepackage{algorithm}
\usepackage{algpseudocode}
\usepackage{algorithmicx}
\usepackage[hang,flushmargin]{footmisc}
\usepackage[accsupp]{axessibility}  % Improves PDF readability for those with disabilities.
\definecolor{iccvblue}{rgb}{0.21,0.49,0.74}
\usepackage[pagebackref,breaklinks,colorlinks,allcolors=iccvblue]{hyperref}
\usepackage{makecell}

\def\paperID{5573} % *** Enter the Paper ID here
\def\confName{ICCV}
\def\confYear{2025}

\title{Steering Guidance for Personalized Text-to-Image Diffusion Models}

\newcommand\blfootnote[1]{%
  \begingroup
  \renewcommand\thefootnote{}\footnote{#1}%
  \addtocounter{footnote}{-1}%
  \endgroup
}
\author{
Sunghyun Park$^{\ast}$\quad Seokeon Choi$^{\ast}$ \quad Hyoungwoo Park\quad Sungrack Yun\\
Qualcomm AI Research$^\dagger$\\
\texttt{\footnotesize\{sunpar, seokchoi, hwoopark, sungrack\}@qti.qualcomm.com}
}

\begin{document}

\makeatletter
\g@addto@macro\@maketitle{
    \centering
    \vspace{-0.5cm}
    \includegraphics[width=0.99\textwidth]{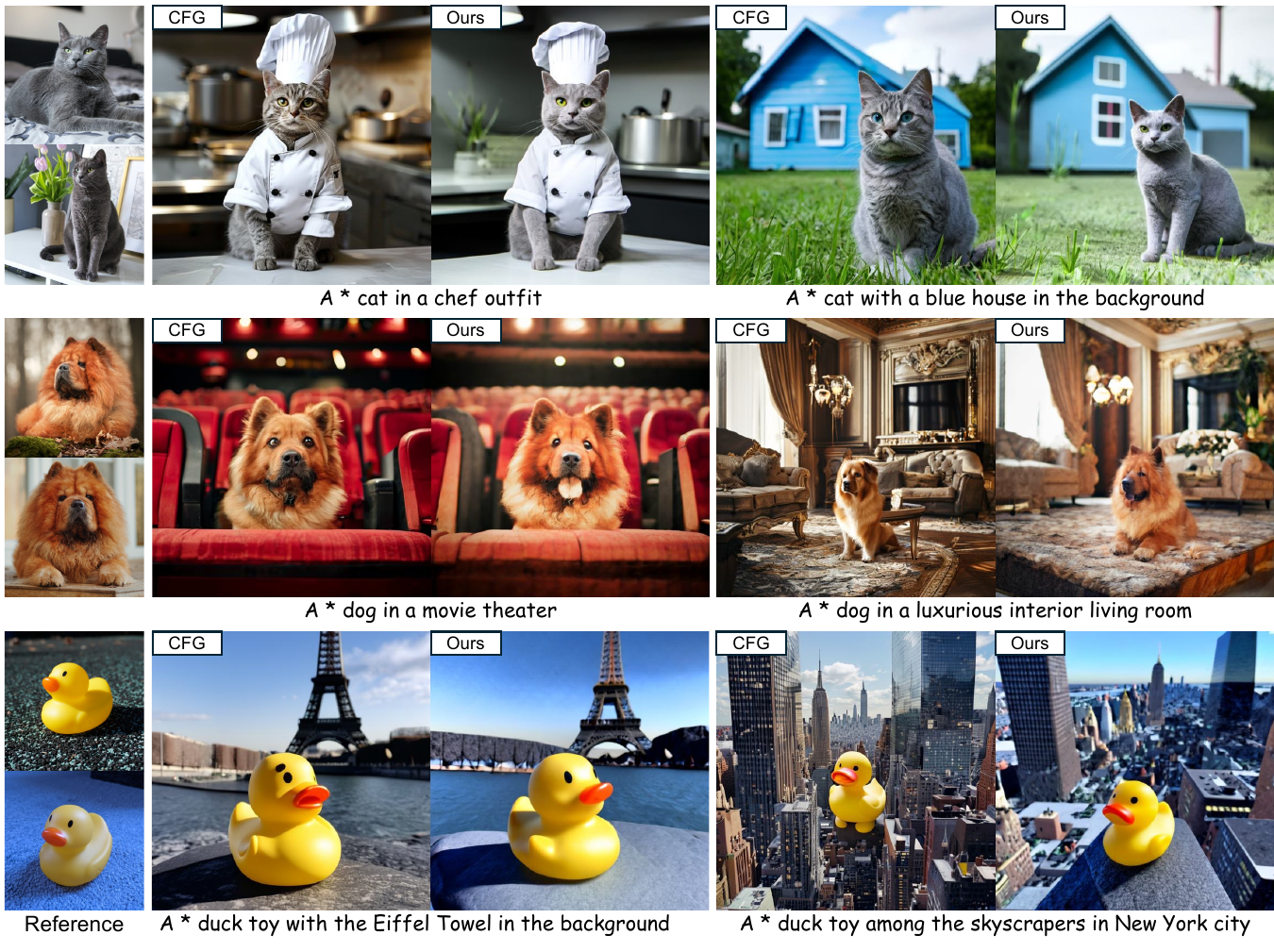}
    \vspace{-0.4cm}
    \captionof{figure}{Comparison of generated image quality between Classifier-Free Guidance (CFG) and our Personalization Guidance.}
    \label{fig:teaser}
    \vspace{0.3cm}
}
\maketitle

%% New version %%
\blfootnote{$^*$These two authors contributed equally to this work.}
\blfootnote{$^\dagger$Qualcomm AI Research is an initiative of Qualcomm Technologies, Inc.}

\begin{abstract}
    Personalizing text-to-image diffusion models is crucial for adapting the pre-trained models to specific target concepts, enabling diverse image generation.
    However, fine-tuning with few images introduces an inherent trade-off between aligning with the target distribution (e.g., subject fidelity) and preserving the broad knowledge of the original model (e.g., text editability).
    Existing sampling guidance methods, such as classifier-free guidance (CFG) and autoguidance (AG), fail to effectively guide the output toward well-balanced space: CFG restricts the adaptation to the target distribution, while AG compromises text alignment. 
    To address these limitations, we propose personalization guidance, a simple yet effective method leveraging an unlearned weak model conditioned on a null text prompt.
    Moreover, our method dynamically controls the extent of unlearning in a weak model through weight interpolation between pre-trained and fine-tuned models during inference.
    Unlike existing guidance methods, which depend solely on guidance scales, our method explicitly steers the outputs toward a balanced latent space without additional computational overhead. 
    Experimental results demonstrate that our proposed guidance can improve text alignment and target distribution fidelity, integrating seamlessly with various fine-tuning strategies.
\end{abstract}

% Personalizing text-to-image diffusion models has become essential to customize the pre-trained models to the target concepts, enabling them to generate diverse images of the concepts.
% However, fine-tuning introduces an inherent trade-off: aligning with the target distribution often comes at the cost of catastrophic forgetting, reducing text editability and generalization.
% Existing sampling guidance methods, such as Classifier-Free Guidance (CFG) and AutoGuidance (AG), struggle to balance adaptation to the target distribution and text alignment in fine-tuned models — CFG restricts target adaptation, while AG lacks text editability. 
% We propose a simple yet effective guidance technique that employs an unlearned weak model with a null text prompt, enhancing text alignment without additional inference cost. 
% Furthermore, we introduce weight interpolation between pre-trained and fine-tuned diffusion models for weak model, enabling dynamic control over the trade-off between specialization and generalization. 
% Through extensive experiments, we demonstrate that our approach outperforms existing guidance techniques in fine-tuned diffusion models, achieving better alignment with target distributions while preserving text editability.
\section{Introduction}

% Necessity Personalization Task
Diffusion models~\cite{ddpm,ddim} have emerged as powerful generative models capable of representing complex data distributions, driving significant advancements in high-fidelity image generation~\cite{ldm,imagen,sdxl,sana}.
However, real-world use cases often demand personalized generation, where the model generates not only generic concepts but also specific concepts like a novel character or person~\cite{dreambooth,textual_inversion,customdiffusion}.
This personalization technique is particularly useful for applications such as portrait editing and specific content creation.

% Personalization Methods
Many approaches have been proposed to personalize diffusion models for novel concepts. 
One line of work~\cite{instantbooth,elite,hyperdreambooth,suti} learns a general-purpose encoder capable of adapting to new concepts without test-time finetuning.
However, this approach requires large-scale text-image datasets and considerable computational resources, and must often be re-trained from scratch for each pre-trained model, making it expensive to adapt to the growing variety of text-to-image or text-to-video diffusion architectures. 
An alternative solution is to directly fine-tune a pre-trained diffusion model on the target images, which can be roughly categorized into fine-tuning low-rank adapters~\cite{lora,para}, text embeddings~\cite{textual_inversion}, specific layers~\cite{classdiffusion,customdiffusion}, or the full model parameters~\cite{dreambooth}. 
This strategy can be applied to any pre-trained diffusion model and only requires a few target images, proving effective for personalization in various scenarios.

% Limitations & Challenges of Personalization Method
Despite these advancements, optimization-based personalization remains a nontrivial challenge.
When fine-tuning a model to a specific, often much smaller, target data distribution, there is an inevitable tradeoff between aligning with the target distribution (\eg, learning a novel concept with a few target images) and preserving broad knowledge of the original model (\eg, preserving text editability).
This trade-off manifests in various ways: a fine-tuned model for personalization may generate high-quality domain-specific images but lose text editability, making it difficult to generalize across different prompts. 
Conversely, a model that retains broad knowledge from its original training distribution may struggle to fully align with new concepts introduced in fine-tuning.
Existing approaches attempt to mitigate this trade-off by adjusting the number of fine-tuning iterations~\cite{textual_inversion,dreambooth}, leveraging regularization~\cite{dreambooth,classdiffusion}, reducing the model parameters of updates~\cite{lora,para}, yet they primarily operate at the fine-tuning stage.
Moreover, balancing the trade-off between comprehensive text editability and precise alignment with a new target distribution remains an open problem.

\begin{figure}[t!]
    \centering
    \includegraphics[width=1.0\linewidth]{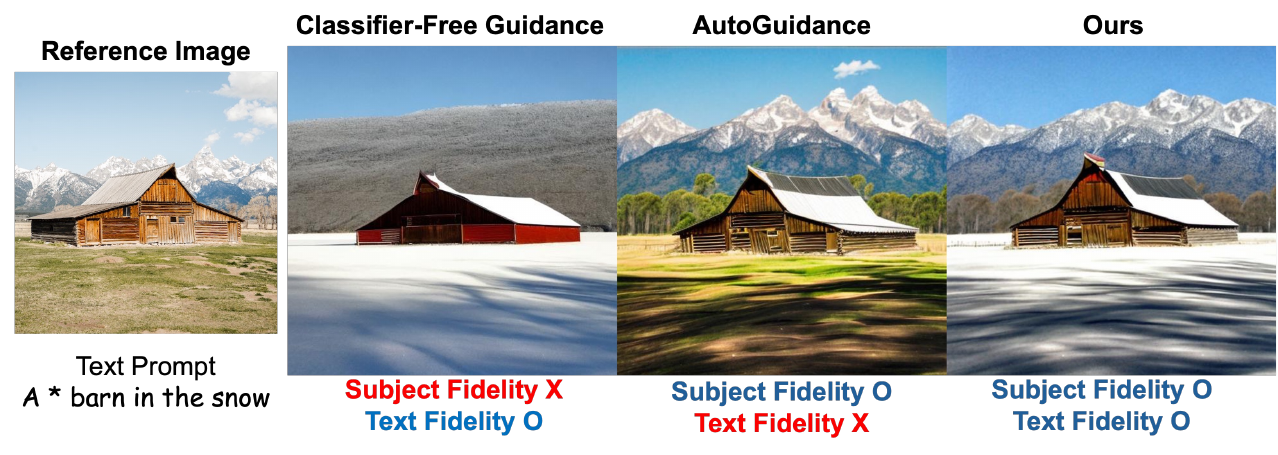}
    \vspace{-0.8cm}
    \caption{Motivation of Personalization Guidance. Classifier-free guidance~\cite{cfg} can interfere with subject fidelity due to its alignment with the text prompt. In contrast, autoguidance~\cite{autoguidance} fails to reflect the text prompt but generates the subject more similarly. Our guidance complements the limitations of both methods.}
    \vspace{-0.6cm}
    \label{fig:motivation}    
\end{figure}

% Guidance
A parallel line of research focuses on sampling guidance techniques, such as Classifier-Free Guidance (CFG)~\cite{cfg} and AutoGuidance (AG)~\cite{autoguidance}. 
These methods rely on “weak models” to detect and push the outputs away from poor trajectories, thereby guiding samples toward higher-quality regions of the data manifold. 
For instance, CFG uses an unconditional model using the null text prompt as the weak model, while AG uses an unlearned model---one with fewer parameters or less training---conditioned on the same text prompt.
However, these methods have limitations when applied to personalized diffusion models, as shown in Fig.~\ref{fig:motivation}. 
CFG can interfere with the model’s adaptation to the target distribution, restricting its ability to fully utilize fine-tuned knowledge. 
Moreover, while the guidance scale of CFG adjusts the degree of text fidelity, it does not improve alignment with the target data.
On the other hand, AG, which primarily enhances alignment with the target distribution, however, lacks an explicit mechanism for improving text alignment, making it unsuitable for fine-tuned text-to-image models.

% Our Method
To address these issues, we propose a simple yet effective guidance technique, called \textbf{Personalization Guidance}, for personalized diffusion models, which can dynamically balance text alignment and adaptation to the target distribution.
Our method leverages an “unlearned” weak model with a null text prompt, without incurring additional computational overhead during inference.
Here, we aim to balance the alignment with the target distribution (\eg, subject fidelity) and the preservation of the broad knowledge of the original model (\eg, text fidelity) by controlling the extent of “unlearning” in a weak model.
Here, the degree of unlearning refers to the extent to which the target data distribution has not been learned compared to the fine-tuned model.
The rationale behind adjusting this degree of unlearning is to steer the output toward an optimal, well-balanced latent space.
Specifically, during inference, we interpolate the weight parameters between the pre-trained and fine-tuned diffusion models to derive an optimal weak model, effectively controlling the extent of unlearning.
Unlike guidance scales, which adjust the intensity of guidance, weight interpolation for the weak model serves to steer outputs in an appropriate direction. 
This straightforward approach can be readily integrated into various fine-tuning diffusion models.

% Our contributions can be summarized as follows:
% \begin{itemize}
%     \item We propose a simple yet effective guidance for personalization, which improves adaptation to the target distribution while preserving knowledge of original models.
%     \item To easily steer the guidance, we use weight interpolation the pre-trained and fine-tuned models, 
%     We leverage weight interpolation between the base and fine-tuned models to dynamically control the trade-off between generalization and specialization.
%     \item Through extensive experiments, we demonstrate that our method outperforms existing guidance techniques in personalized diffusion models.
% \end{itemize}

\section{Related Work}

\noindent\textbf{Personalization of Diffusion Models.}
Personalizing diffusion models aim to generate novel images by adapting to specific concepts using a set of reference images.
Optimization-based methods personalize models by fine-tuning textual embeddings~\cite{textual_inversion,customdiffusion,classdiffusion}, entire model parameters with regularization~\cite{dreambooth}, or low-rank adapters~\cite{lora,para}.
Another line of research leverages extensive text-image datasets to train dedicated text-to-image personalization models, by fine-tuning additional modules~\cite{elite,instantbooth,hyperdreambooth,suti}.
Recently, several studies~\cite{dreammatcher,sag,freeu} have achieved better personalization without additional fine-tuning or training.
Orthogonal to these existing approaches, we introduce a simple yet effective personalization guidance technique that seamlessly integrates with various personalization methods by modifying the classifier-free guidance.
Moreover, our guidance enables us to adjust the degree of subject and text fidelity without incurring additional computational costs.

\noindent\textbf{Guidance for Diffusion Models.}
Classifier-Free Guidance (CFG)~\cite{cfg} enhances conditional image generation by leveraging the unconditional model as a weak model to guide the outputs toward better adherence to the given condition.
Various guidance techniques have been explored, such as the guidance using perturbations~\cite{pag,stg} or blurring to the attention maps~\cite{selfattentionguidance}.
Autoguidance~\cite{autoguidance} improves generation by using a weak model---one with fewer parameters or less training---to push outputs toward a well-trained direction.
While InstructPix2Pix~\cite{instructpix2pix} introduces separate CFG using both image and text conditions, it cannot be directly applied to DB-LoRA due to its reliance on image-conditioned training.
For personalization, subject-agnostic guidance~\cite{sag} enhances subject-agnostic attributes by employing a weak model conditioned on subject-specific text prompts. 
% In this study, we aim to address the fundamental trade-off in personalization-aligning a model with a smaller target data distribution while preserving the broad text fidelity of the original model by deriving a weak model that dynamically adjusts the degree of adaptation, balancing comprehensive text editability with precise alignment to new concepts.
In this study, we aim to address the fundamental personalization trade‑off---aligning a model with a smaller target data distribution while preserving the broad text fidelity of the original model. We derive a weak model that dynamically adjusts the degree of adaptation, balancing comprehensive text editability with precise alignment to new concepts.
% In this study, our objective is to enhance the alignment between the original diffusion model's knowledge (\ie text fidelity) and the target data by deriving a weak model that modulates the degree of adaptation to the target data.

\noindent\textbf{Weight Interpolation.}
There are previous studies on exploring the use of weight interpolation.
WISE-FT~\cite{wiseft} improves the robustness of fine-tuned classifiers against distribution shifts through weight-space ensembling. 
Modelsoups~\cite{modelsoups} maximizes model accuracy and robustness by interpolating the weights of multiple fine-tuned classifiers trained with different hyperparameter configurations.

\section{Method}
\label{sec:method}

\subsection{Preliminaries}
\label{sec:preliminaries}

\noindent\textbf{Diffusion Models.}
Diffusion models~\cite{ddpm,ddim} aim to learn reversing a forward noising process that gradually transforms data samples into noise. 
Concretely, the forward process starts with a data distribution $p_0(\mathbf{x})$ and ends at a high-noise distribution $p_T(\mathbf{x}) \approx \mathcal{N}(\mathbf{0}, \mathbf{I})$. 
Sampling from the data distribution proceeds by solving the equivalent probability-flow ordinary differential equation (PF-ODE). 
Under the variance-preserving parameterization where $p(\mathbf{x}_t|\mathbf{x}_0)=\mathcal{N}(\mathbf{x}_0,\,\sigma_t^2\mathbf{I})$, the PF-ODE takes the form:
\begin{equation}
    \mathrm{d}\mathbf{x}_t = - \sigma_t\,\nabla_{\mathbf{x}_t}\!\log p(\mathbf{x}_t)\,\mathrm{d}t,
    \quad
    \mathbf{x}_T \sim p_T(\mathbf{x}_T).
\end{equation}
In diffusion models, the score function is parameterized by $\epsilon_\theta$.
Here, the noise estimation of the diffusion model can be considered as $\epsilon_\theta(\mathbf{x}_t) \approx - \sigma_t\,\nabla_{\mathbf{x}_t}\!\log p(\mathbf{x}_t)$.

\noindent\textbf{Personalization.}
Suppose we have a pre-trained diffusion model $\epsilon_{\theta}$, where $\theta$ is the parameters of the model.
We wish to personalize it to a smaller target dataset $\mathcal{D}_\text{target} = \{(\mathbf{x}^i, \mathbf{c}^i)\}_{i=1}^N$, where each $\mathbf{x}^i$ and $\mathbf{c}^i$ is an image and its associated text prompt, respectively.
The goal is to obtain new parameters $\theta'$ so that the fine-tuned model $\epsilon_{\theta'}$ can generate samples more faithfully matched the $\mathcal{D}_\text{target}$.

Fine-tuning typically follows the same denoising objective used by the original diffusion model~\cite{ddpm}.
Concretely, let $\sigma$ be a noise level sampled from a training schedule, and let $\boldsymbol{\epsilon}\sim \mathcal{N}(\mathbf{0}, \sigma^2 \mathbf{I})$. 
For each $(\mathbf{x}, \mathbf{c})$ in $\mathcal{D}_\text{target}$, we form the noisy input $\mathbf{x}_t = \mathbf{x} + \boldsymbol{\epsilon}$ and minimize:
\begin{equation}
    \label{eq:finetune_obj}
    \mathcal{L} = \mathbb{E}_{(\mathbf{x},\mathbf{c})\sim \mathcal{D}_\text{target},\, \sigma,\, \boldsymbol{\epsilon}}
    \Bigl\| \epsilon_{\theta'}(\mathbf{x}_t, \sigma, \mathbf{c}) - \boldsymbol{\epsilon} \Bigr\|^2_2.
\end{equation}
This teaches the model to reconstruct $\mathbf{x}$ from noisy inputs in the domain of $\mathcal{D}_\text{target}$, thereby adapting it to the new concept.
% Here, $\mathcal{L}(\theta') \leq \mathcal{L(\theta)}$.
In practice, additional regularization terms are often used to prevent overfitting~\cite{dreambooth, classdiffusion}.

\noindent\textbf{Classifier-Free Guidance.}
To enhance the generation towards an arbitrary condition $\mathbf{c}$, classifier-free guidance (CFG)~\cite{cfg} introduces a new sampling distribution $p^\lambda(\mathbf{x}|\mathbf{c})$ composed with both $p(\mathbf{x})$ and the classifier distribution $p(c|\mathbf{x})^\lambda$, which is described as:
\begin{equation}
    p^\lambda(\mathbf{x}|\mathbf{c}) \propto p(\mathbf{x})\ p(\mathbf{c}|\mathbf{x})^\lambda,
\end{equation}
where $\lambda>1$ amplifies the conditional likelihood.
By applying Bayes’ rule for some timestep $t$, the corresponding score function is:
\begin{multline}
    \nabla_{\mathbf{x}_t} \log p^\lambda(\mathbf{x}_t|\mathbf{c}) =
    \nabla_{\mathbf{x}_t} \log p(\mathbf{x}_t) \\
    + \lambda \bigl(\nabla_{\mathbf{x}_t} \log p(\mathbf{x}_t | \mathbf{c}) - \nabla_{\mathbf{x}_t} \log p(\mathbf{x}_t) \bigr).
\end{multline}
To enable conditional generation such as text prompts and class labels, CFG jointly trains the unconditional model $\epsilon_\theta(\mathbf{x}_t, \phi)$ and the conditional model $\epsilon_\theta(\mathbf{x}_t, \mathbf{c})$ within a single model by dropping $\mathbf{c}$ with certain probability to allow null conditioning with $c=\phi$.
In the personalization setting, parameterizing the score function with the fine-tuned diffusion model $\epsilon_{\theta'}$ is described as:
\begin{equation}
    \tilde\epsilon^{\lambda}_{\theta'}(\mathbf{x}_t | \mathbf{c}) = \textcolor{cyan}{\epsilon_{\theta'}(\mathbf{x}_t | \phi)} + \lambda \bigl( \epsilon_{\theta'}(\mathbf{x}_t | \mathbf{c}) - \textcolor{cyan}{\epsilon_{\theta'}(\mathbf{x}_t | \phi)} \bigr).
\end{equation}
However, while CFG improves alignment with the conditions, such as the text prompt, it struggles to enhance adaptation to the target data in the personalization task.

\begin{figure}[t!]
    \centering
    \includegraphics[width=1.0\linewidth]{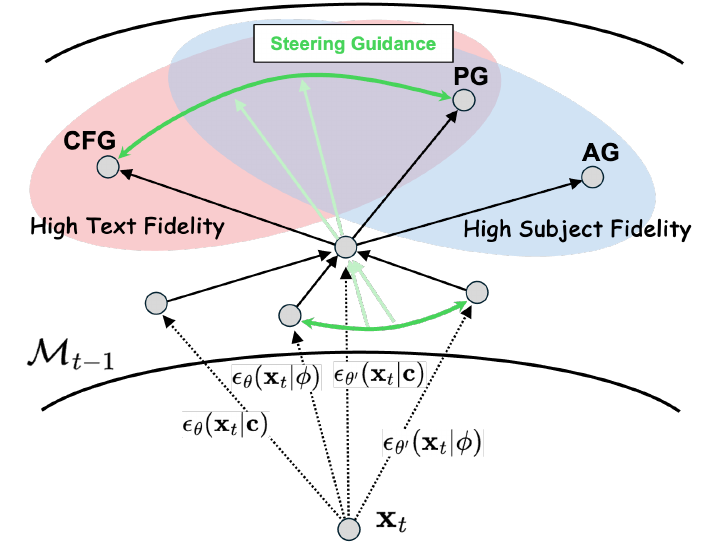}
    \vspace{-0.5cm}
    \caption{Comparison between Classifier-Free Guidance (CFG), AutoGuidance (AG), and our Personalization Guidance (PG), with the band conceptually representing the noisy data manifold. By leveraging weight interpolation, guidance can be steered toward the optimal space between CFG and PG.}
    \vspace{-0.4cm}
    \label{fig:method}
\end{figure}

\noindent\textbf{AutoGuidance.}
AutoGuidance (AG)~\cite{autoguidance} is another guidance technique that employs a weaker version of the same model to identify suboptimal generation trajectories.
Rather than using the unconditional model $\epsilon_\theta(\mathbf{x}_t | \phi)$, AG diretly guides a model with the weak model $\epsilon_{\theta_{weak}}$, which is trained on the same task and conditioning, but degraded by under-training or less-parameters.
However, obtaining a weaker version $\epsilon_{\theta_{weak}}$ of a fully trained diffusion model like Stable Diffusion~\cite{ldm} is challenging in practice.
Instead, we found that a less trained model could serve as the pre-trained model for personalization, where $\theta_{weak}=\theta$, effectively shifts the output away in the direction of fine-tuning, enhancing the adaptation to the target data. 
In the personalization, AG is formulated as:
\begin{equation}
    \tilde\epsilon^{\lambda}_{\theta'}(\mathbf{x}_t | \mathbf{c}) = \textcolor{orange}{\epsilon_{\theta}(\mathbf{x}_t | \mathbf{c})} + \lambda \bigl( \epsilon_{\theta'}(\mathbf{x}_t | \mathbf{c}) - \textcolor{orange}{\epsilon_{\theta}(\mathbf{x}_t | \mathbf{c})} \bigr).
\end{equation}
However, this approach has a major drawback---it scarcely reflects the text prompt, as shown in Fig.~\ref{fig:motivation}.

\begin{figure}[t!]
    \centering
    \includegraphics[width=1.0\linewidth]{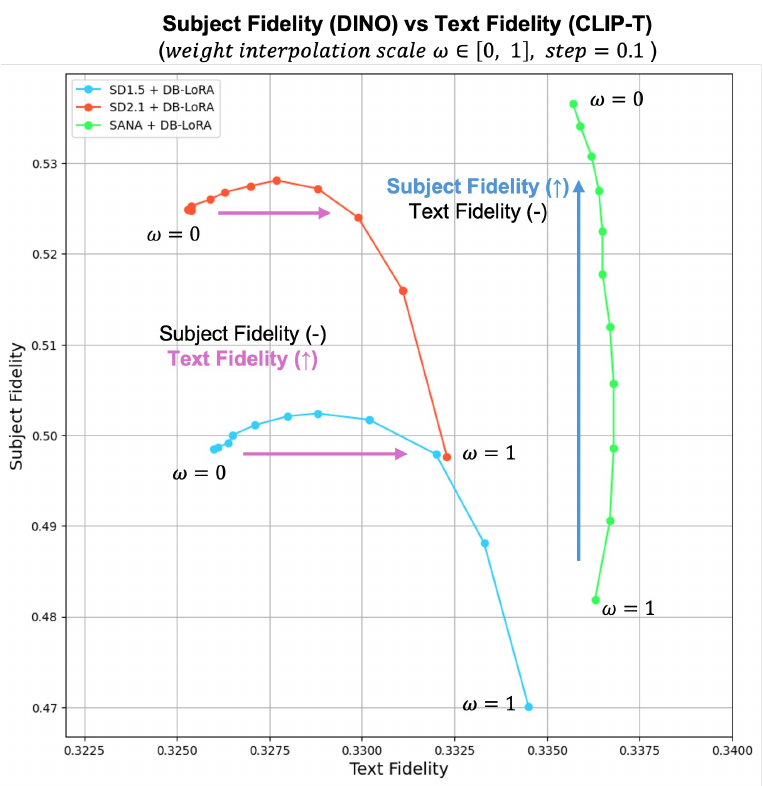}
    \vspace{-0.7cm}
    \caption{Ablation study on the weight interpolation scale $\omega \in [0.0, 1.0]$ with measurements taken at intervals of 0.1. Here, we use DreamBooth (DB) LoRA based on SD 1.5, SD 2.1, and SANA. By adjusting $\omega$, it is possible to improve subject fidelity or text fidelity, while preserving the other.}
    \vspace{-0.4cm}
    \label{fig:ablation_graph_omega}
\end{figure}

\subsection{Steering Personalization Guidance}

% 수식 서술할게 많아짐.
% \begin{algorithm}[t!]
% \small
% \caption{Conditional Sampling with Steered Personalization Guidance}
% \label{alg:cfg}
% \begin{algorithmic}[1]
% \Require $\mathbf{x}_T \sim \mathcal{N}(0, I), 0 \leq \lambda \in \mathbb{R}$
% % \KwIn{$\mathbf{x}_T \sim \mathcal{N}(0, I), 0 \leq \lambda \in \mathbb{R}$}
% \For{$i=T$ {\bfseries to} $1$}
%     \State{$\tilde\epsilon^\lambda_{\theta'}(\mathbf{x}_t|\mathbf{c}) = \epsilon_{\theta_\omega}(\mathbf{x}_t|\phi) + \lambda \bigl(\epsilon_{\theta'}(\mathbf{x}_t|\mathbf{c}) - \epsilon_{\theta_\omega}(\mathbf{x}_t|\phi)\bigr)$}
%     \State{$\mathbf{x}_{t-1} \sim \mathcal{N} \bigl( \frac{1}{\sqrt{\alpha_t}} \left( \mathbf{x}_t - \frac{1-\alpha_t}{\sqrt{1-\bar{\alpha}_t}} \tilde\epsilon^\lambda_{\theta'}(\mathbf{x}_t|\mathbf{c}) \right), \Sigma_t \bigr)$}
%     % \State{$\mathbf{x}(\mathbf{x}_t) \gets (\mathbf{x}_t - \sqrt{1-\bar\alpha_t}\epsilon(\mathbf{x}_t))/\sqrt{\bar\alpha_t}$}
%     % \State{$\mathbf{x}_{t-1} = \sqrt{\bar\alpha_{t-1}} (\mathbf{x}_t) + \sqrt{1-\bar\alpha_{t-1}}\epsilon(\mathbf{x}_t)$}
%     \EndFor
%     \State {\bfseries return} $\mathbf{x}_0$
% \end{algorithmic}
% \end{algorithm}

Inspired by autoguidance~\cite{autoguidance}, we simply modify the CFG~\cite{cfg}, by leveraging a pre-trained diffusion model $\epsilon_{\theta}$ instead of the fine-tuned model $\epsilon_{\theta'}$ as the weak model.
\begin{equation}
    \tilde\epsilon^{\lambda}_{\theta'}(\mathbf{x}_t|\mathbf{c}) = \textcolor{magenta}{\epsilon_{\theta}(\mathbf{x}_t | \phi)} + \lambda \bigl( \epsilon_{\theta'}(\mathbf{x}_t | \mathbf{c}) - \textcolor{magenta}{\epsilon_{\theta}(\mathbf{x}_t | \phi)} \bigr).
\end{equation}
Despite being a simple extension, we found that this approach significantly improves subject fidelity.

Fig.~\ref{fig:method} shows the rationale behind our personalization guidance.
Since this simply modifed guidance can steer the output toward the intersection space where both text fidelity and subject fidelity are high.
As illustrated in Fig.~\ref{fig:method}, we explore the feasibility of identifying a superior space within the noisy data manifold. 
We hypothesize that an optimal point for text and subject fidelity could be located between CFG and PG, and we propose a method to facilitate the steering of guidance. 
Steering the guidance involves the adjustment of an appropriate weak model. 
By identifying a suitable weak model, we can steer $\epsilon_{\theta}(\mathbf{x}_t | \phi)$ towards a more favorable point. 
We propose employing a weight interpolation to obtain an appropriate weak model.
Specifically, by interpolating weights between the parameters of the pre-trained and the fine-tuned models, we can modulate the degree of adaptation to the target data, thereby obtaining an appropriate weak model $\epsilon_{\theta_{\omega}}$.
\begin{equation}
    \theta_\omega = \omega \,\theta' + (1 - \omega)\,\theta, \quad \omega \in [0,1],
\end{equation}
where $\omega$ denotes the scale of weight interpolation.
Increasing $\omega$ moves parameters closer to the fine-tuned model $\theta'$, reducing the loss $\mathcal{L}$ on $\mathcal{D}_{\text{target}}$.
% Interpolating weights as $\theta_\omega$ smoothly moves from original scores towards fine-tuned parameters
Finally, applying the weight interpolation to the weak model in the guidance:
\begin{equation}
    \tilde\epsilon^{\lambda}_{\theta'}(\mathbf{x}_t|\mathbf{c}) = \textcolor{magenta}{\epsilon_{\theta_{\omega}}(\mathbf{x}_t | \phi)} + \lambda \bigl( \epsilon_{\theta'}(\mathbf{x}_t | \mathbf{c}) - \textcolor{magenta}{\epsilon_{\theta_{\omega}}(\mathbf{x}_t | \phi)} \bigr).
\end{equation}
Note that $\omega=1$ is equal to CFG.
As shown in Fig.~\ref{fig:ablation_graph_omega}, it is possible to easily control the degree of subject and text fidelity.
If $\theta$ includes only linear classifier, the personalization guidance with weight interpolation is equal to the combination of CFG and AG, which is the output-space interpolation.
However, as neural networks are non-linear with respect to parameters, guidance with weight interpolation is different from the output interpolation of CFG and AG~\cite{wiseft}.

% TODO: 이론적인 근거 & 수식 더 추가해보기 / Algorithm 작성하기 (CFG와 비교)

% Why weight interpolation for weak model can 
% Hypothesis 1. linear model connectivity (LMC) w.r.t. a single performance measure (e.g., loss) . 
% LMC holds if all losses for the interpolated weights exceed the interpolated outputs..? -> 이거
% 

\section{Experiments}

\begin{table}[t!]
\centering
\setlength{\tabcolsep}{5pt} % Reduce column padding
\small
\begin{tabular}{c||l|c|ccc}
    \toprule
    & \multicolumn{1}{c|}{\textbf{Method}} & \textbf{$\lambda$} & \textbf{DINO} & \textbf{CLIP-I} & \textbf{CLIP-T} \\
    \midrule
    & DB-LoRA & - & 0.3741 & 0.6797 & 0.2834 \\
    & + CFG & 7.5 & 0.4701 & 0.7349 & \textbf{0.3345} \\
    & \quad + AG & 2.0 & 0.4831 & 0.7408 & 0.3236 \\
    \rowcolor{gray!10} \cellcolor{white} 
    & + Ours ($\omega=0$) & 7.5 & \underline{0.4985} & \textbf{\underline{0.7516}} & 0.3260 \\
    \rowcolor{gray!10} \cellcolor{white} 
    \multirow{-5}{*}{\rotatebox{90}{\textbf{SD 1.5}}}
    & + Ours ($\omega=0.6$) & 7.5 & \textbf{\underline{0.5024}} & \underline{0.7510} & \underline{0.3288} \\
    \midrule
    & DB-LoRA & - & 0.4202 & 0.7076 & 0.2929 \\
    & + CFG & 7.5 & 0.4976 & 0.7519 & \textbf{\underline{0.3323}} \\
    & \quad + AG & 2.0 & 0.5072 & 0.7571 & 0.3212 \\
    \rowcolor{gray!10} \cellcolor{white} 
    & + Ours ($\omega=0$) & 7.5 & \underline{0.5248} & \underline{0.7655} & 0.3254 \\
    \rowcolor{gray!10} \cellcolor{white} 
    \multirow{-5}{*}{\rotatebox{90}{\textbf{SD 2.1}}}
    & + Ours ($\omega=0.6$) & 7.5 & \textbf{\underline{0.5281}} & \textbf{\underline{0.7660}} & \underline{0.3277} \\
    \midrule
    \multirow{4}{*}{\rotatebox{90}{\textbf{SANA}}}
    & DB-LoRA & - & 0.4190 & 0.6939 & 0.3064 \\
    & + CFG & 4.5 & \underline{0.4819} & \underline{0.7291} & \textbf{\underline{0.3363}} \\
    & \quad + AG & 2.0 & 0.4792 & 0.7284 & 0.3355 \\
    \rowcolor{gray!10} \cellcolor{white} 
    & + Ours ($\omega=0$) & 4.5 & \textbf{\underline{0.5366}} & \textbf{\underline{0.7522}} & \underline{0.3357} \\
    % \rowcolor{gray!10} \cellcolor{white} 
    % & \quad + Steer & 4.5 &  \\
    \bottomrule
\end{tabular}
\vspace{-0.3cm}
\caption{Comparison with other guidance methods across different base diffusion models including SD1.5, SD2.1, and SANA. Here, we use \textbf{DreamBooth-LoRA (DB-LoRA)} only.}
\vspace{-0.5cm}
\label{table:DB-LoRA}
\end{table}

\subsection{Experimental Setting}

% In this section, we evaluate our proposed approach for diffusion-based personalization on the ViCo dataset. We detail the dataset, evaluation metrics, implementation setup, and baselines, followed by quantitative and qualitative results.

\noindent\textbf{Datasets.}
% ViCo Dataset
We conduct our experiments on the ViCo dataset~\cite{vico}, which includes 16 concepts and 31 text prompts.
Each concept includes a small set of reference images (typically 4-7) and associated text prompts.
Following the previoius work~\cite{dreammatcher}, we fine-tune the pre-trained text-to-image diffusion models per concept and then generate 8 images per concept and text prompt, resulting in a total of 3,968 images for evaluation.
% The detailed procedure and the prompt list are described in the supplementary.

\noindent\textbf{Evaluation Metrics.}
% Personalization: CLIP-I, CLIP-T, DINO scores
% Human Preference Optimization: HPS-v2, PickScore, Aesthetics, CLIP-T
% In-Context LoRA: Qualitative Results Only?
% + User Study
Following the existing personalization studies~\cite{dreambooth,dreammatcher,classdiffusion}, we evaluate subject and text fidelity.
For the quantitative assessment, we adopt three evaluation metrics:
(i) CLIP-T score~\cite{clip}, which uses CLIP’s text-to-image similarity to evaluate text prompt alignment;
(ii) CLIP-I score, which computes image-to-image similarity using CLIP embeddings to gauge how closely generated samples match reference images;  and 
(iii) DINO score~\cite{dino}, which calculates image-to-image similarity between reference and generated images, offering a complementary perspective on subject fidelity. 

\begin{table}[t!]
\centering
\setlength{\tabcolsep}{5pt} % Reduce column padding
\small
\begin{tabular}{c||l|c|ccc}
    \toprule
    & \multicolumn{1}{c|}{\textbf{Method}} & \textbf{$\lambda$} & \textbf{DINO} & \textbf{CLIP-I} & \textbf{CLIP-T} \\
    \midrule
    & DB-LoRA+TI & - & 0.3725 & 0.6775 & 0.2817 \\
    & + CFG & 7.5 & 0.4618 & 0.7292 & \textbf{\underline{0.3316}} \\
    & \quad + SAG & 1.0 & 0.4600 & 0.7273 & \underline{0.3307} \\
    & \quad + AG & 2.0 & 0.4806 & 0.7408 & 0.3264 \\
    \rowcolor{gray!10} \cellcolor{white} 
    & + Ours ($\omega=0$) & 7.5 & \underline{0.4814} & \underline{0.7459} & 0.3233 \\    
    \rowcolor{gray!10} \cellcolor{white} 
    & + Ours ($\omega=0.7$) & 7.5 & \textbf{\underline{0.4880}} & \textbf{\underline{0.7461}} & 0.3272 \\
    \cmidrule{2-6}
    & ClassDiffusion & - & 0.3703 & 0.6669 & 0.2757 \\
    & + CFG & 7.5 & 0.5287 & 0.7517 & \textbf{\underline{0.3305}} \\
    & \quad + SAG & 1.0 & 0.5172 & 0.7465 & 0.3296 \\
    & \quad + AG & 2.0 & 0.5570 & 0.7628 & 0.3280 \\
    \rowcolor{gray!10} \cellcolor{white} 
    \multirow{-11}{*}{\rotatebox{90}{\textbf{SD 1.5}}}
    & + Ours ($\omega=0$) & 7.5 & \textbf{\underline{0.5700}} & \textbf{\underline{0.7656}} & \underline{0.3297} \\ 
    \midrule
    & DB-LoRA+TI & - & 0.4514 & 0.7251 & 0.2823 \\
    & + CFG & 7.5 & 0.5291 & 0.7749 & \textbf{\underline{0.3244}} \\
    & \quad + SAG & 1.0 & 0.5328 & 0.7766 & \underline{0.3223} \\
    & \quad + AG & 2.0 & 0.5557 & 0.7877 & 0.3206 \\
    \rowcolor{gray!10} \cellcolor{white} 
    & + Ours ($\omega=0$) & 7.5 & \underline{0.5683} & \underline{0.7918} & 0.3163 \\
    \rowcolor{gray!10} \cellcolor{white} 
    & + Ours ($\omega=0.6$) & 7.5 & \textbf{\underline{0.5689}} & \textbf{\underline{0.7919}} & 0.3185\\
    \cmidrule{2-6}
    & ClassDiffusion & - & 0.4273 & 0.6995 & 0.2843 \\
    & + CFG & 7.5 & 0.5840 & 0.7757 & \textbf{\underline{0.3277}} \\
    & \quad + SAG & 1.0 & 0.5783 & 0.7739 & 0.3244 \\
    & \quad + AG & 2.0 & 0.6060 & 0.7856 & \underline{0.3246} \\
    \rowcolor{gray!10} \cellcolor{white} 
    \multirow{-11}{*}{\rotatebox{90}{\textbf{SD 2.1}}}
    & + Ours ($\omega=0$) & 7.5 & \textbf{\underline{0.6166}} & \textbf{\underline{0.7896}} & \underline{0.3246} \\ 
    \bottomrule
\end{tabular}
\vspace{-0.3cm}
\caption{Comparison with other guidance methods across different base diffusion models including SD1.5 and SD2.1. Here, we use \textbf{DreamBooth-LoRA (DB-LoRA) + Textual Inversion (TI)} and \textbf{ClassDiffusion}. Since these personalization methods fine-tune text embedding, subject-agnostic guidance (SAG) is also compared with our method.}
\vspace{-0.5cm}
\label{table:others}
\end{table}

\begin{figure*}[t!]
    \centering
    \includegraphics[width=1.0\linewidth]{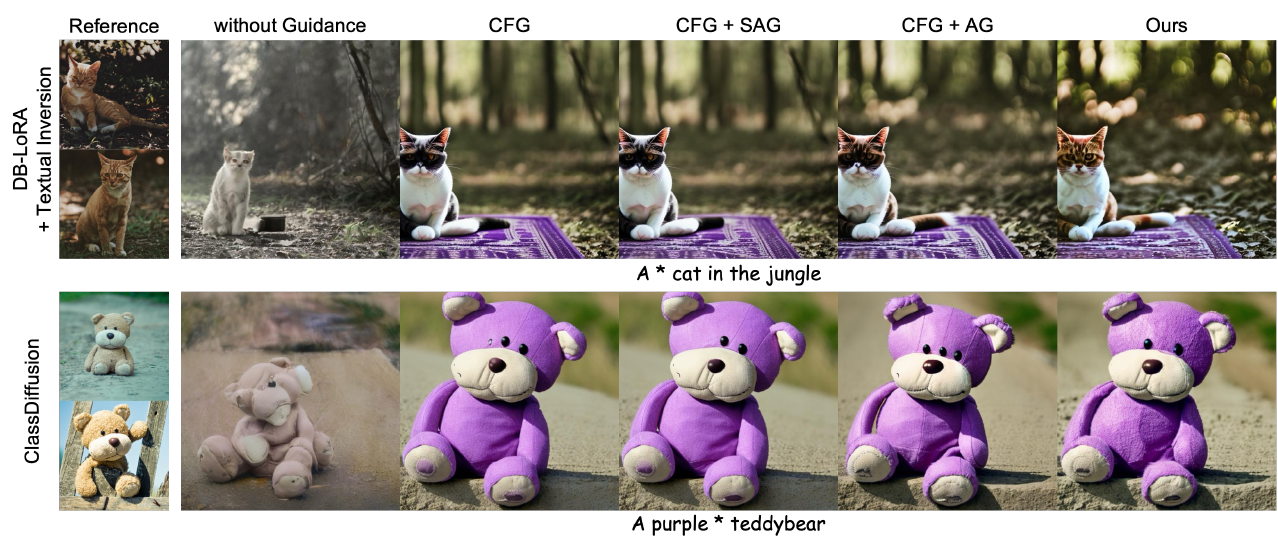}
    \vspace{-0.8cm}
    \caption{Comparison with other guidance techniques, including without guidance, CFG, CFG+SAG, and CFG+AG. These images are generated by the fine-tuned SD 2.1 using DB-LoRA and ClassDiffusion.}
    \vspace{-0.2cm}
    \label{fig:comparison_sd21}
\end{figure*}

\begin{figure*}[t!]
    \centering
    \includegraphics[width=1.0\linewidth]{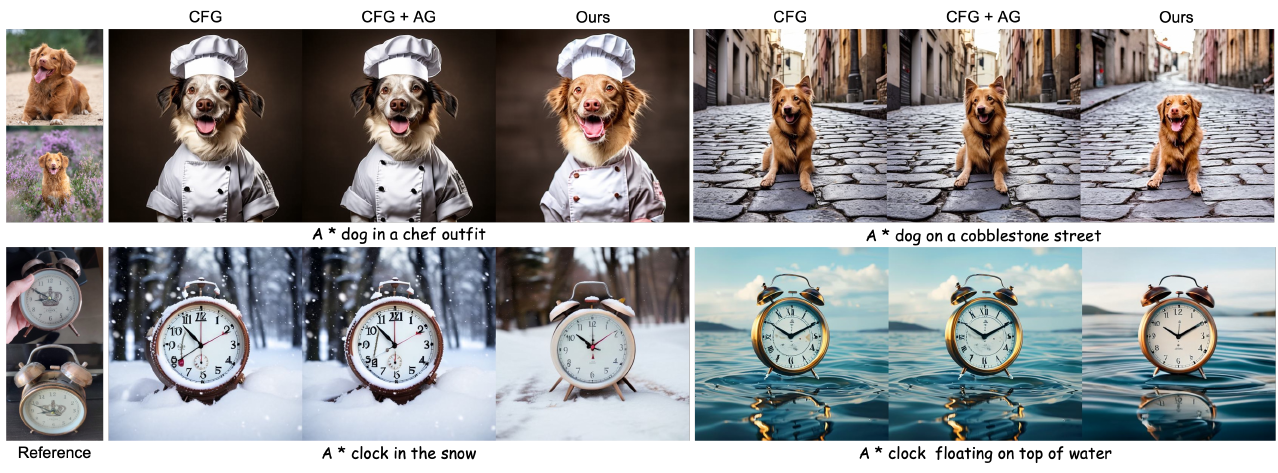}
    \vspace{-0.8cm}
    \caption{Comparison with other guidance techniques. These images are generated by fine-tuned SANA~\cite{sana} using DB-LoRA.}
    \vspace{-0.2cm}
    \label{fig:comparison_sana}
\end{figure*}

\noindent\textbf{Implementation Details.}
% Model List: SD1.5, SD2.1, FLUX (?), RealisticVision, Diffusion-DPO
% Personalization Methods: DreamBooth, DB-LoRA, DB-LoRA + Textual Inversion, ClassDiffusion
In our experiments, we utilize three pre-trained text-to-image diffusion models as the base models: Stable Diffusion 1.5 (SD 1.5)~\cite{ldm}, Stable Diffusion 2.1 (SD 2.1)~\cite{ldm}, and SANA-1.6B~\cite{sana}.
SD 1.5 and SD 2.1 generate 512$\times$512 resolution images, which comprise U-Net based architectures.
On the other hand, SANA is a recent text-to-image diffusion model, which is based on the diffusion transformer~\cite{dit} that can generate images up to 4096$\times$4096 resolution.
In the experiments, SANA generates 1024$\times$1024 images.

To fine-tune the diffusion models, we leverage several personalization techniques: (i) DreamBoooth-LoRA (DB-LoRA)~\cite{dreambooth,lora}, (ii) DB-LoRA + Textual Inversion (TI), and (iii) ClassDiffusion~\cite{classdiffusion}.
We fine-tune each model for up to 500 steps on the reference images from the ViCo dataset.
We set the number of sampling steps (\eg, DDIM) between 20-50.
Further details regarding additional hyperparameters are described in the supplementary.

\noindent\textbf{Baselines.}
% Guidance: Classifier-Free Guidance, AutoGuidance, Classifier-Free Guidance ++, Subject-Agnostic Guidance
%        ㄴ 추가한다면: Perturbed Attention Guidance, Self-Attention Guidance (?), AutoLoRA (?)
We evaluate our guidance method against other guidance techniques: (i) Classifier-Free Guidance (CFG)~\cite{cfg}, (ii) Subject-Agnostic Guidance (SAG)~\cite{sag}, and (iii) AutoGuidance (AG)~\cite{autoguidance}.
% SAG와 AG는 혼자서는 text prompt를 잘 반영해서 이미지를 생성해내는게 어렵기 때문에, CFG와 같이 사용해서 이미지를 생성한다.
% 특히, AG는 최근 CFG와 AG를 같이 쓰는 work을 참고하여 실험을 하였다.
Since SAG and AG alone struggle to generate images that accurately reflect the text prompt, they are used in conjunction with CFG to produce images. 
Specifically, AG experiments are conducted by referring to recent work~\cite{autolora} that combines CFG and AG.

\subsection{Results}

\noindent\textbf{Quantitative Results.}
Table~\ref{table:DB-LoRA} and Table~\ref{table:others} showcase the personalization performance on the ViCo dataset. 
Table~\ref{table:DB-LoRA} provides the quantitative results for DB-LoRA based on SD 1.5, SD 2.1, and SANA. 
Since DB-LoRA+TI and ClassDiffusion involve fine-tuning textual embeddings, we also evaluate subject-agnostic guidance (SAG)~\cite{sag}, which necessitates learned textual embeddings for specific concepts. 
As illustrated in both Table~\ref{table:DB-LoRA} and Table~\ref{table:others}, our guidance method significantly enhances subject fidelity (DINO and CLIP-I scores), while maintaining text fidelity (CLIP-T score) with minimal change. 
Additionally, by adjusting $\omega$, the model can further optimize personalization performance across all evaluation metrics.
We select the optimal $\omega$ based on the best DINO and CLIP-I scores. 
For SANA with DB-LoRA and ClassDiffusion, the best subject fidelity was observed at $\omega=0$.
These results demonstrate that our approach significantly improves subject fidelity while maintaining text fidelity, and that steering the guidance enables a balanced trade-off between the two.

Interestingly, the impact of personalized guidance varies depending on the personalization method. 
ClassDiffusion~\cite{classdiffusion} employs a semantic preservation loss that maintains the text embedding space, leading to better overall text fidelity retention. 
As a result, our guidance method maintains the CLIP-T score more effectively when applied to ClassDiffusion than to other personalization methods. 
This observation highlights a key insight: the better the fine-tuning direction, the more effective our guidance becomes.

\begin{figure*}[t!]
    \centering
    \includegraphics[width=0.9\linewidth]{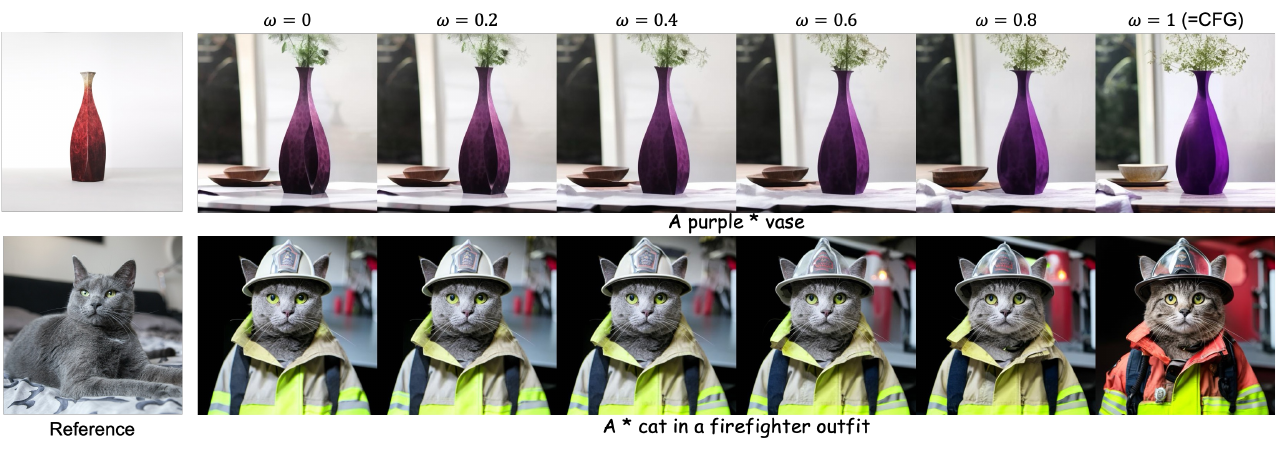}
    \vspace{-0.5cm}
    \caption{Generated images by changing $\omega$, where SANA with DB-LoRA is used.}
    \vspace{-0.3cm}
    \label{fig:ablation_omega}
\end{figure*}

\noindent\textbf{Qualitative Results.}
Fig.~\ref{fig:comparison_sd21} and Fig.~\ref{fig:comparison_sana} show the comparison results based on SD 2.1 and SANA, respectively. 
As shown in Fig.~\ref{fig:comparison_sd21}, CFG often fails to capture fine details of the reference subject (\eg, cat's appearance and the teddybear's eye).
Additionally, applying SAG or AG has minimal impact in some cases (first row of Fig.~\ref{fig:comparison_sd21}) or even degrades image quality, introducing artifacts such as a teddy bear with three eyes (CFG+AG, second row of Fig.~\ref{fig:comparison_sd21}).
In contrast, our method consistently achieves better subject fidelity with the same noise input.
Similarly, in Fig.~\ref{fig:comparison_sana}, our method better preserves subject details compared to CFG and CFG+AG, as evidenced by the significant improvement in subject fidelity in quantitative results.
For instance, it accurately retains the ears and color of a dog (first row of Fig.~\ref{fig:comparison_sana}) and the shape and numerical details of a clock (second row of Fig.~\ref{fig:comparison_sana}). 
In contrast, applying AG with CFG shows little to no improvement despite incurring a higher computational cost compared to both CFG and our guidance method.

\begin{table}[t!]
\centering
\small
\begin{tabular}{l|cc}
    \toprule
    \multicolumn{1}{c|}{\textbf{Method}} & \textbf{Subject Fidelity $\uparrow$} & \textbf{Text Fidelity $\uparrow$} \\
    \midrule
    \multicolumn{3}{l}{\textbf{Stable Diffusion 2.1 + ClassDiffusion}} \\
    \midrule
    CFG         & 10.45\% & 15.24\% \\
    CFG + SAG   & 8.01\% & 6.20\% \\
    CFG + AG    & 12.60\% & 6.20\% \\
    Ours        & \textbf{55.22\%} & 32.04\% \\
    Undecided   & 13.67\% & \textbf{40.31\%} \\
    \midrule
    \multicolumn{3}{l}{\textbf{SANA + DB-LoRA}} \\
    \midrule
    CFG         & 2.48\% & 3.08\% \\
    CFG + AG    & 2.17\% & 2.77\% \\
    Ours        & \textbf{68.32\%} & 8.64\% \\
    Undecided   & 27.01\% & \textbf{85.49\%} \\    
    \bottomrule
\end{tabular}
\vspace{-0.2cm}
\caption{User preference on subject fidelity and text fidelity.}
\vspace{-0.6cm}
\label{table:userstudy}
\end{table}

\noindent\textbf{User Study.}
Table~\ref{table:userstudy} presents the user preference results comparing our method with baseline approaches. 
The results show that our guidance technique achieves significantly higher subject fidelity than other guidance methods. 
Our approach also shows superior text fidelity in user evaluations. 
Further details can be found in the supplementary.

\begin{figure}[t!]
    \centering
    \vspace{-0.2cm}
    \includegraphics[width=1.0\linewidth]{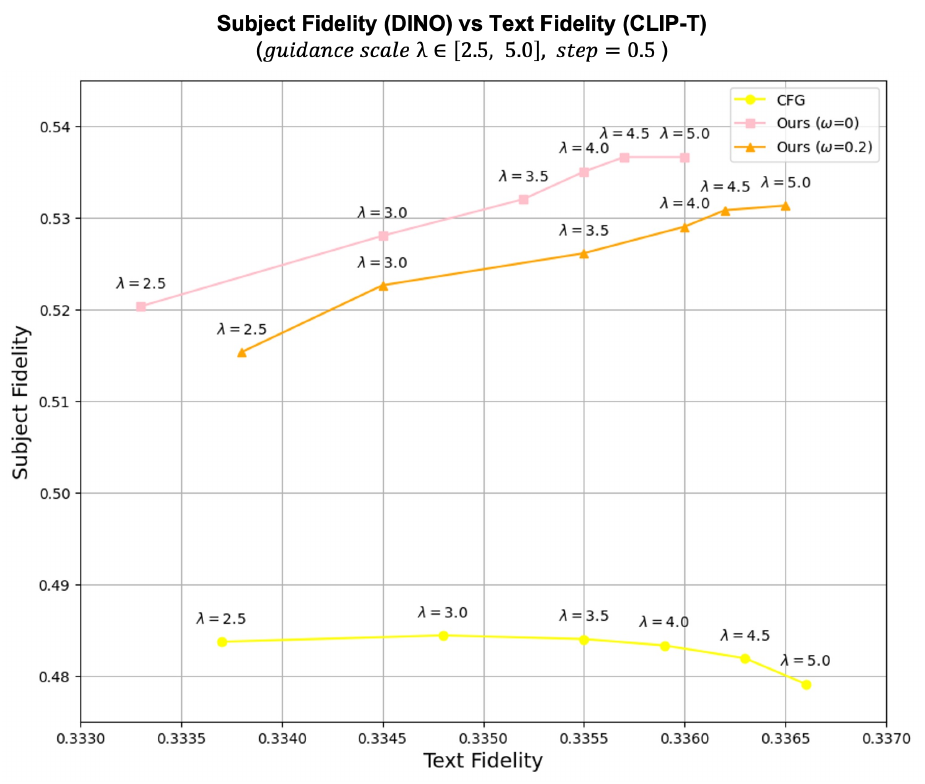}
    \vspace{-0.8cm}
    \caption{Ablation study on the guidance scale $\lambda \in [2.5, 5.0]$, measuring performance at 0.5 intervals using DB-LoRA based on SANA. When $\omega$ is set to 0.2, we can improve both subject and text fidelity compared to CFG, regardless of guidance scale $\lambda$. Note that CLIP-T score is slightly changed in the graph ($max(\text{CLIP-T}) - min(\text{CLIP-T}) < 0.004$).}
    \vspace{-0.7cm}
    \label{fig:ablation_graph_lambda}
\end{figure}

\subsection{Further Analysis}

\noindent\textbf{Ablation Study on $\omega$.}
Fig.~\ref{fig:ablation_graph_omega} illustrates how DINO and CLIP-T scores vary with changes in $\omega$.
For both SD 1.5 and SD 2.1, personalization guidance slightly reduces text fidelity but significantly enhances subject fidelity compared to CFG ($\omega=1$).
By appropriately adjusting $\omega$, it is possible to either boost text fidelity while maintaining subject fidelity (as indicated by the pink arrow) or enhance subject fidelity while preserving text fidelity (as indicated by the blue arrow).
It is important to note that these trends may vary depending on the personalization technique or the pre-trained diffusion model.
However, the ability to easily fine-tune results via weight interpolation at inference, without modifying training steps or learning rates, offers a major practical advantage over traditional methods for controlling overfitting.
Fig.~\ref{fig:ablation_omega} further visualizes the impact of $\omega$ on generation quality.
The first row shows that higher $\omega$ values retain fine details, such as the intricate patterns on the vase, at the cost of weaker adherence to the "purple" text prompt. 
Additionally, as $\omega$ decreases, subject fidelity improves significantly, particularly in the range of 0.6 to 1.0.

\noindent\textbf{Ablation Study on Guidance Scale.}
Fig.~\ref{fig:ablation_graph_lambda} shows how personalization performance varies with different guidance scales $\lambda$.
We evaluate our guidance at $\omega=0$ and $\omega=0.2$, where text fidelity remains preserved.
As $\lambda$ decreases, text fidelity deteriorates for both guidance methods. 
However, as $\lambda$ increases, our guidance method shows improvements in both the DINO score and CLIP-T score, whereas the DINO score of CFG declines.
This suggests that CFG reduces subject fidelity, and adjusting the guidance scale alone is insufficient to correct this effect.

\begin{table}[t!]
\centering
\setlength{\tabcolsep}{3pt} % Reduce column padding
\vspace{-0.2cm}
\begin{minipage}[t]{0.45\linewidth}
\centering
\resizebox{0.92\linewidth}{!}{%
\begin{tabular}{l|c}
    \toprule
    \multicolumn{1}{c|}{\textbf{Method}} & \textbf{PickScore$\uparrow$} \\
    \midrule
    DPO+CFG & 21.53 \\
    DPO+Ours & \textbf{21.58} \\
    \bottomrule
\end{tabular}}
\vspace{-0.2cm}
\caption{Results of combining Diffusion-DPO with ours.}
\vspace{-0.4cm}
\label{table:diffusion-dpo-score}
\end{minipage}
\hfill
\begin{minipage}[t]{0.53\linewidth}
\centering
\resizebox{1.0\linewidth}{!}{%
\begin{tabular}{l|c}
    \toprule
    \multicolumn{1}{c|}{\textbf{Method}} & \textbf{DreamSim$\downarrow$} \\
    \midrule
    PairCustom+CFG  & 0.7993 \\
    PairCustom+Ours & \textbf{0.7344} \\ % w=0
    \bottomrule
\end{tabular}}
\vspace{-0.2cm}
\caption{Results of combining PairCustom with ours.}
\vspace{-0.4cm}
\label{table:paircustom-score}
\end{minipage}
\end{table}

\begin{figure}
    \centering
    \includegraphics[width=\linewidth]{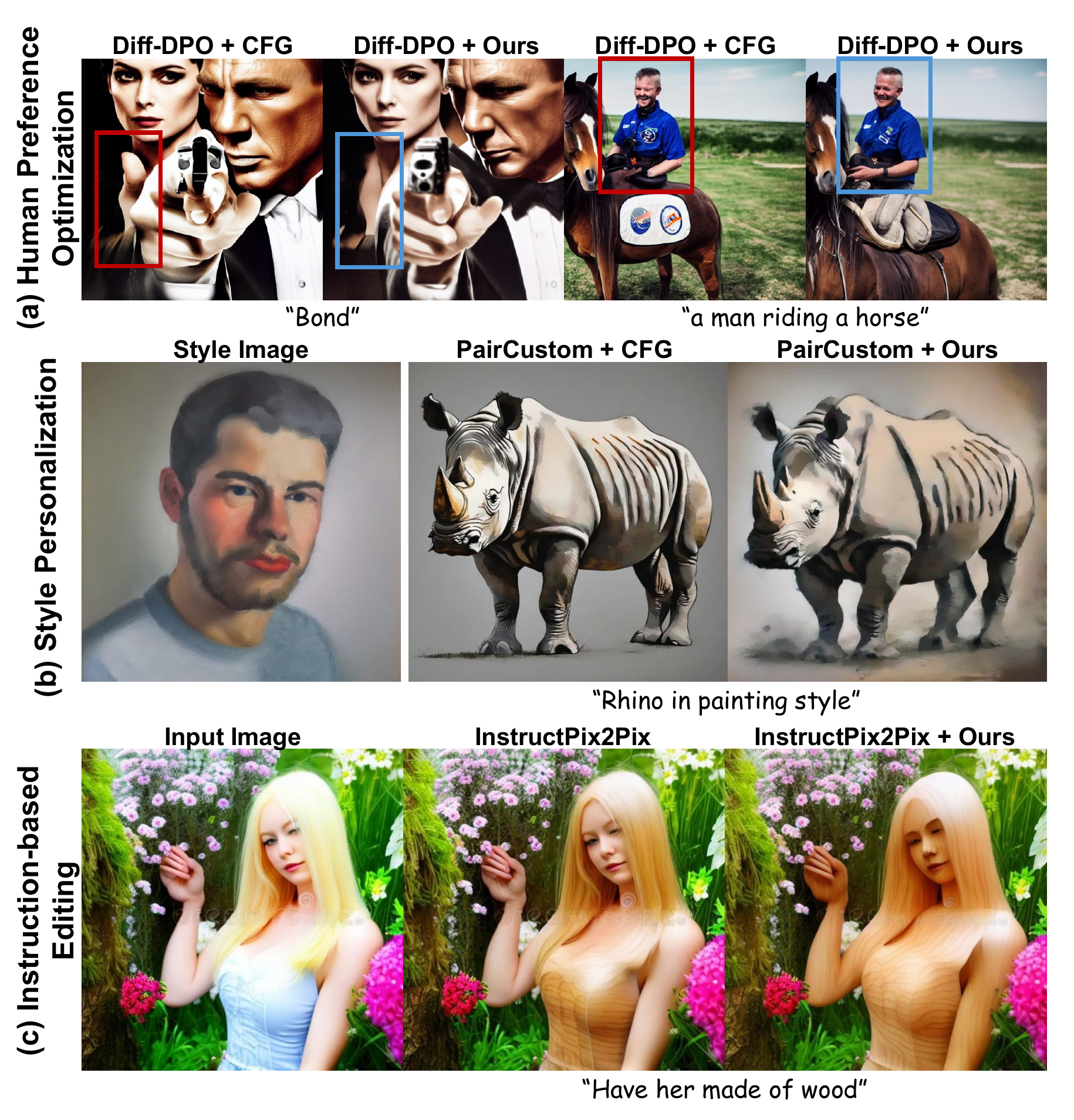}
    \vspace{-0.8cm}
    \caption{Qualitative results of our guidance across diverse tasks.}
    \vspace{-0.5cm}
    \label{fig:application}
\end{figure}

\subsection{Application to Fine-tuned Diffusion Models.}

Fine-tuning diffusion model extends beyond personalization tasks to applications such as style personalization~\cite{paircustom}, human preference learning~\cite{diffusiondpo,diffusionkto}, and specific task learning~\cite{iclora,instructpix2pix}. 
Our personalization guidance framework leverages the knowledge gap between the pre-trained and fine-tuned models, emphasizing the information better captured by the fine-tuned model. 
This concept can be generalized to control the degree of adaptation for other fine-tuned models, such as adjusting the extent of human preference alignment or style personalization during inference. 

To this end, to investigate our method's generalizability across diverse tasks, Fig.~\ref{fig:application} show the following experiments: (a) Human Preference Optimization (Diffusion-DPO~\cite{diffusiondpo}), (b) Style Personalization (PairCustom~\cite{paircustom}), and (c) Instruction-based Editing (InstructPix2Pix~\cite{instructpix2pix}).

\noindent\textbf{(a) Human Preference Optimization}: For evaluation with the Diffusion-DPO model~\cite{diffusiondpo}, we use the PartiPrompts dataset~\cite{parti} and report PickScore~\cite{parti}, which reflects human-perceived image quality.
Table~\ref{table:diffusion-dpo-score} shows that integrating our method into the SD 1.5-based Diffusion-DPO model yields superior performance compared to CFG, confirming the effectiveness of our approach on human preference optimization tasks. 
As shown in Fig.~\ref{fig:application} (a), our method helps generate more natural-looking images and corrects artifacts that would otherwise appear unnatural (\textit{e.g.}, additional finger).

\noindent\textbf{(b) Style Personalization}: 
To evaluate style alignment, we use two pre-trained style-LoRAs released by PairCustom, combining each with both CFG and our method. 
Using text prompts from the StyleAlign~\cite{stylealign} paper, we generate 800 images and compute the DreamSim score~\cite{dreamsim} to quantify the similarity between the target style and generated outputs. 
As shown in Table~\ref{table:paircustom-score}, applying our method significantly improves style alignment, demonstrating its effectiveness for style personalization.
% PairCustom 정성 결과도 설명
In Fig.~\ref{fig:application} (b), combining PairCustom with our guidance leads to a more faithful transfer of the painting style from the reference image.
% \textbf{(c) Encoder-based Customization}: 
% Fig.~\ref{fig:application} (c) presents qualitative result using our method in integration with IP-Adapter~\cite{ipadapter}, an encoder-based customization approach.
% Compared to CFG, our method produces images that more accurately preserve the facial identity from the reference image, highlighting its compatibility with encoder-based pipelines.

\noindent\textbf{(c) Instruction-based Editing}:
In Fig.~\ref{fig:application} (c), our guidance improve the performance of the InstructPix2Pix~\cite{instructpix2pix}, which is designed to edit images based on text instructions, to better follow the given instruction. 
As a result, the generated outputs are more accurately aligned with the intended edits.

These results demonstrate that applying our guidance to various fine-tuned diffusion models allows them to generate images more faithfully aligned with their respective fine-tuning objectives.
This highlights the generalizability of our method across diverse tasks.
Further implementation details are provided in the supplementary.

\section{Conclusion}

This paper tackles the critical challenge of balancing adaptation to the target distribution and knowledge preservation from the original model in personalized diffusion models.
Our approach simply modifies the existing guidance by adjusting the degree of unlearning in the weak model through weight interpolation between the pre-trained and fine-tuned models.
This simple yet effective guidance enhances subject fidelity while preserving text editability, ensuring an optimal trajectory for personalization.
Furthermore, we demonstrate that our guidance can be effectively applied to a wide range of fine-tuned diffusion models, such as human preference learning.
% Our findings open new research directions, with potential applications in various fine-tuned diffusion models, including personalization, detoxification, and human preference learning.

\newpage

\noindent\textbf{Acknowledgement}
We would like to thank the Qualcomm AI Research team for their valuable discussions.

{
    \small
    \bibliographystyle{ieeenat_fullname}
    \bibliography{reference}
}

\clearpage
\setcounter{page}{1}
\setcounter{section}{0}
\renewcommand{\thesection}{\Alph{section}}
\maketitlesupplementary

\section{Implementation Details}

\noindent\textbf{Hyperparameter Details.}
% Following the Diffusers' DreamBooth implementation code, we standardized the guidance scale to 7.5 across all main experiments. 
Fine-tuning was conducted with a batch size of 1 over 500 iterations (except for ClassDiffusion~\cite{classdiffusion}). 
For experiments involving Stable Diffusion versions 1.5 and 2.1~\cite{ldm}, both training and inference utilized images at a 512×512 resolution, whereas SANA~\cite{sana} employed a 1024×1024 resolution. 
In addition, the maximum diffusion timestep is set to 1,000.
During inference, the DDIM scheduler~\cite{ddim} was applied to Stable Diffusion 1.5 and 2.1 with 50 inference steps. 
Conversely, SANA followed its original implementation by employing the Flow-DPM-Solver~\cite{dpmsolver} with 20 inference steps.

For $\omega$ that is a scale for weight interpolation, a trade-off exists between subject and text fidelity.  
Thus, selecting optimal $\omega$ should align with the intended objective: values in the range of 0–0.3 are preferable when maximizing subject fidelity, while values around 0.4–0.6 offer a balanced improvement with better preservation of text fidelity.

\noindent\textbf{Details of Personalization Methods.}
Following the implementation codes provided by Diffusers or the official SANA repository, DreamBooth-LoRA~\cite{dreambooth,lora} was configured with a rank of 4 and a learning rate of 1e-4 with prior preservation loss. 
For the combined DreamBooth-LoRA and Textual Inversion~\cite{textual_inversion} approach, the same rank and learning rate were maintained, with the number of learnable textual embeddings set to 2. 
We utilized the original implementation code of ClassDiffusion~\cite{classdiffusion}, with a training batch size of 2, a learning rate of 1e-5, and applied augmentation to the training images.
Additionally, Table~\ref{table:prompt_list} shows the prompt list we utilized to generate images for evaluation.

\noindent\textbf{Details of Guidances (Baselines).}
For classifier-free guidance~\cite{cfg}, experiments were conducted with the commonly used guidance scale of 7.5. 
In the case of autoguidance~\cite{autoguidance}, based on Stable Diffusion 2.1, we experimented with guidance scales ranging from 2.0 to 5.0 during inference, reporting the best-performing value ($\lambda$ = 2.0). 
Additionally, subject-agnostic guidance~\cite{sag} was re-implemented, referencing the original paper, specifically for methods employing textual inversion.

\noindent\textbf{User Study.}
We assessed user preferences between Stable Diffusion 2.1 combined with ClassDiffusion and SANA with DreamBooth-LoRA. 
Participants were instructed to select images exhibiting the highest subject fidelity and text fidelity. 
If a clear preference was indiscernible across all image results, they could choose 'undecided.' 
Each model was evaluated over 15 sets, totaling 30 sets for user preference assessment. 
The final results were calculated by averaging the preferences across all participants.

\section{Details of Application Experiemnts.}
\noindent\textbf{(a) Diffusion-DPO}
We used a Diffusion-DPO model that was fine-tuned from Stable Diffusion 1.5 (SD 1.5). Since Diffusion-DPO is a fine-tuned version of SD 1.5, we constructed the weak model by performing weight interpolation between SD 1.5 and Diffusion-DPO. The interpolation coefficient $\omega$ was set to 0.4.

\noindent\textbf{(b) PairCustom}
For the PairCustom setting, we used a style LoRA model trained using the official repository. Since LoRA is also used in subject personalization tasks, we adopted the same experimental setup. In this case, we set $\omega = 0.0$, meaning that the weak model corresponds to the base SD 1.5 model.

\noindent\textbf{(c) InstructPix2Pix}
InstructPix2Pix is also a model fine-tuned from SD 1.5 for image editing based on natural language instructions. Therefore, we constructed the weak model by interpolating the weights of InstructPix2Pix and SD 1.5. We set $\omega = 0.0$, which effectively uses SD 1.5 as the weak model. In addition, InstructPix2Pix employs a separate classifier-free guidance (CFG) mechanism. For generating the unconditional output (i.e., using a null text prompt), we used the SD 1.5 model as the weak model.

% 알고리즘 작성?
% 증명..?

\begin{table}[t!]
\centering
\setlength{\tabcolsep}{4.5pt} % Reduce column padding
\small
\begin{tabular}{c||l|c|ccc}
    \toprule
    & \multicolumn{1}{c|}{\textbf{Method}} & \textbf{$\lambda$} & \textbf{DINO} & \textbf{CLIP-I} & \textbf{CLIP-T} \\
    \midrule
    \multirow{5}{*}{\rotatebox{90}{\textbf{SD 1.5}}}
    & DB-LoRA & - & 0.3741 & 0.6797 & 0.2834 \\
    & \quad + CFG & 7.5 & 0.4701 & 0.7349 & \textbf{0.3345} \\    
    & \quad + Ours & 7.5 & \textbf{0.4985} & \textbf{0.7516} & 0.3260 \\
    \cmidrule{2-6}
    & \quad + CFG++ & 0.4 & 0.4738 & 0.7352 & \textbf{0.3321} \\    
    & \quad + Ours++ & 0.4 & \textbf{0.5059} & \textbf{0.7510} & 0.3250 \\
    \midrule
    \multirow{5}{*}{\rotatebox{90}{\textbf{SD 2.1}}}
    & DB-LoRA & - & 0.4202 & 0.7076 & 0.2929 \\
    & \quad + CFG & 0.4 & 0.4976 & 0.7519 & \textbf{0.3323} \\    
    & \quad + Ours & 0.4 & \textbf{0.5248} & \textbf{0.7655} & 0.3254 \\
    \cmidrule{2-6}
    & \quad + CFG++ & 0.4 & 0.5105 & 0.7531 & \textbf{0.3304} \\    
    & \quad + Ours++ & 0.4 & \textbf{0.5385} & \textbf{0.7680} & 0.3245 \\
    \bottomrule
\end{tabular}
\vspace{-0.2cm}
\caption{Integration with CFG++ and our method.}
\label{table:cfgpp}
\end{table}

\begin{table*}[t!]
\centering
\small
\resizebox{0.7\linewidth}{!}{%
\begin{tabular}{l|c|ccc||ccc}
    \toprule
    \multirow{2}{*}{\textbf{Method}} & \multirow{2}{*}{\textbf{Rank}} & \multicolumn{3}{c||}{SD 1.5} & \multicolumn{3}{c}{SD 2.1} \\
    & & \textbf{DINO} & \textbf{CLIP-I} & \textbf{CLIP-T}
    & \textbf{DINO} & \textbf{CLIP-I} & \textbf{CLIP-T} \\
    \midrule
    +CFG  & 8  & 0.4900 & 0.7445 & \textbf{0.3335} & 0.5154 & 0.7571 & \textbf{0.3315} \\
    \rowcolor{gray!10}
    +Ours & 8  & \textbf{0.5211} & \textbf{0.7584} & \textbf{0.3335} & \textbf{0.5396} & \textbf{0.7690} & 0.3280 \\
    \midrule
    +CFG  & 16 & 0.5105 & 0.7519 & \textbf{0.3337} & 0.5288 & 0.7637 & \textbf{0.3302} \\
    \rowcolor{gray!10}
    +Ours & 16 & \textbf{0.5349} & \textbf{0.7633} & \textbf{0.3337} & \textbf{0.5574} & \textbf{0.7781} & 0.3285 \\
    \midrule
    +CFG  & 32 & 0.5468 & 0.7644 & \textbf{0.3309} & 0.5551 & 0.7730 & \textbf{0.3277} \\
    \rowcolor{gray!10}
    +Ours & 32 & \textbf{0.5763} & \textbf{0.7794} & 0.3272 & \textbf{0.5845} & \textbf{0.7872} & 0.3251 \\
    \bottomrule
\end{tabular}}
\vspace{-0.2cm}
\caption{Comparison of DB-LoRA with varying ranks on SD 1.5 and SD 2.1. Our method consistently outperforms CFG in subject fidelity.}
\label{table:ablation_rank}
\end{table*}

\section{Additional Results}

\noindent\textbf{Integration with CFG++~\cite{cfgpp}.}
To address mode collapse issues in classifier-free guidance (CFG), CFG++~\cite{cfgpp} has been introduced as an enhancement to the CFG method.
CFG++ offers a refined approach to guidance in diffusion models, overcoming several drawbacks of traditional CFG and leading to more reliable and higher-quality text-to-image generation.
To demonstrate the applicability of our method, we conducted experiments integrating it with CFG++. 
Table~\ref{table:cfgpp} illustrates that our method significantly enhances subject fidelity not only when combined with CFG but also when integrated with CFG++. In this experiment, we adopted a simple approach by setting $\omega=0$.

\begin{table}[t!]
\centering
\small
\begin{tabular}{l|cc}
    \toprule
    \multicolumn{1}{c|}{\textbf{Method}} & \textbf{Max Memory} &  \textbf{Inference Time} \\
    \midrule
    CFG         & 6,591 MB & 14.5 secs \\
    CFG + SAG   & 6,595 MB & 21.2 secs \\
    CFG + AG    & 6,595 MB & 21.2 secs \\
    Ours        & 6,591 MB & 14.5 secs \\
    \bottomrule
\end{tabular}
\vspace{-0.2cm}
\caption{Comparison of computational cost. Tested using SD 2.1 models with DB-LoRA + TI.}
% \vspace{-0.4cm}
\label{table:computational_cost}
\end{table}

\noindent\textbf{Computational Cost Analysis.}
We evaluate the computational efficiency of our method by reporting the maximum allocated memory and average inference time. As shown in Table~\ref{table:computational_cost}, our approach introduces no additional computational overhead during guidance. All measurements were conducted using an A5000 GPU.

\begin{figure}[h!]
    \centering
    \includegraphics[width=0.93\linewidth]{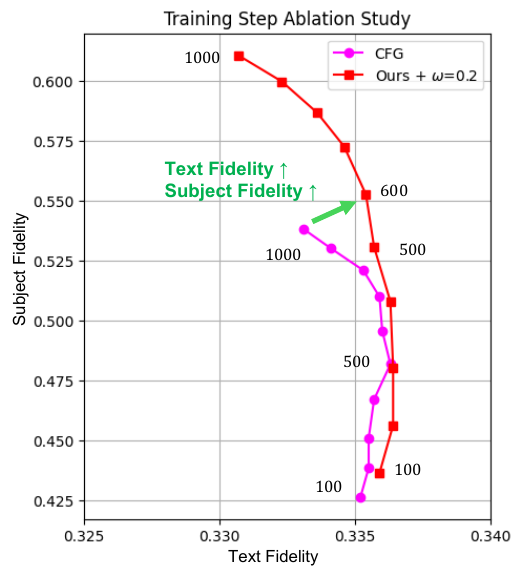}
    \vspace{-0.3cm}
    \caption{Ablation study on the number of training steps. Notably, our method achieves superior subject fidelity with fewer steps, demonstrating enhanced efficiency.}
    % \vspace{-0.3cm}
    \label{fig:ablation_graph_trainingstep}
\end{figure}

\noindent\textbf{Further Study on DB-LoRA.}
We conducted additional experiments to enhance subject fidelity by increasing the number of training steps and the LoRA rank.
Fig.~\ref{fig:ablation_graph_trainingstep} presents the results of DB-LoRA based on SANA with varying numbers of training steps. 
Notably, DB-LoRA with Ours for 600 steps outperforms DB-LoRA with CFG for 1000 steps in both subject and text fidelity, highlighting the superior efficiency and effectiveness of our approach.
Table~\ref{table:ablation_rank} compares DB-LoRA based on SD 1.5 and SD 2.1 with varying LoRA ranks. 
Across both models, our method outperforms CFG in terms of subject fidelity. 
In some cases, even with half the rank (\textit{i.e.}, fewer parameters), our method achieves better subject fidelity.

\noindent\textbf{Detailed Results of Ablation Study on $\omega$.}
As shown in Fig.~\ref{fig:ablation_graph_omega}, we summarize the performance variations in subject fidelity and text fidelity based on different $\omega$ values in Table~\ref{table:omega_ablation_sd15},~\ref{table:omega_ablation_sd21},~\ref{table:omega_ablation_sana}. 
While the graph visualizes only the DINO score, we also include the CLIP-I score in the table for a more comprehensive evaluation.
In addition, we report the results on DB-LoRA+TI in Table~\ref{table:omega_ablation_sd15_ti},~\ref{table:omega_ablation_sd21_ti}.

\begin{figure*}[t!]
    \centering
    \includegraphics[width=0.95\linewidth]{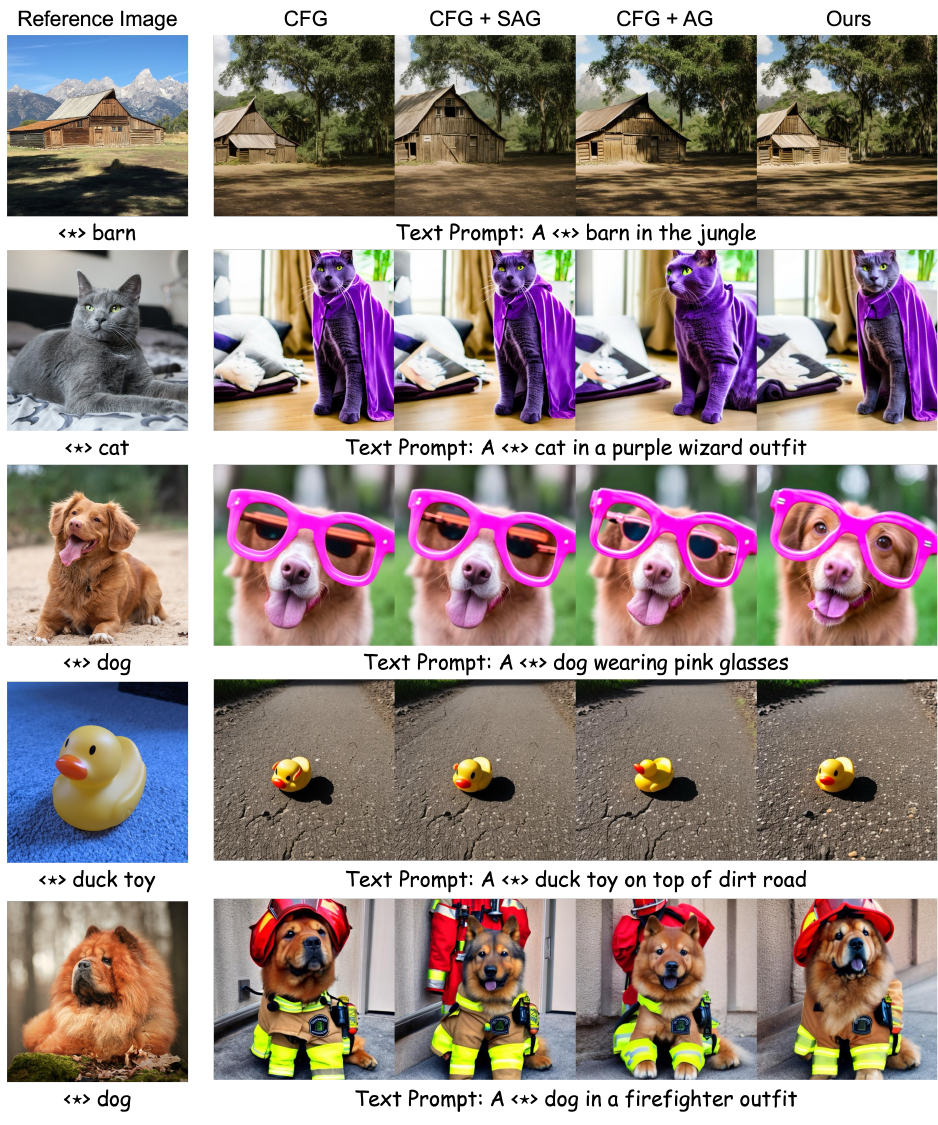}
    % \vspace{-0.5cm}
    \caption{Comparison with other guidance techniques. These images are generated by fine-tuned SD 2.1~\cite{ldm} using ClassDiffusion.}
    % \vspace{-0.2cm}
    \label{supp-fig:comparison_classdiffusion}
\end{figure*}

\begin{figure*}[t!]
    \centering
    \includegraphics[width=0.8\linewidth]{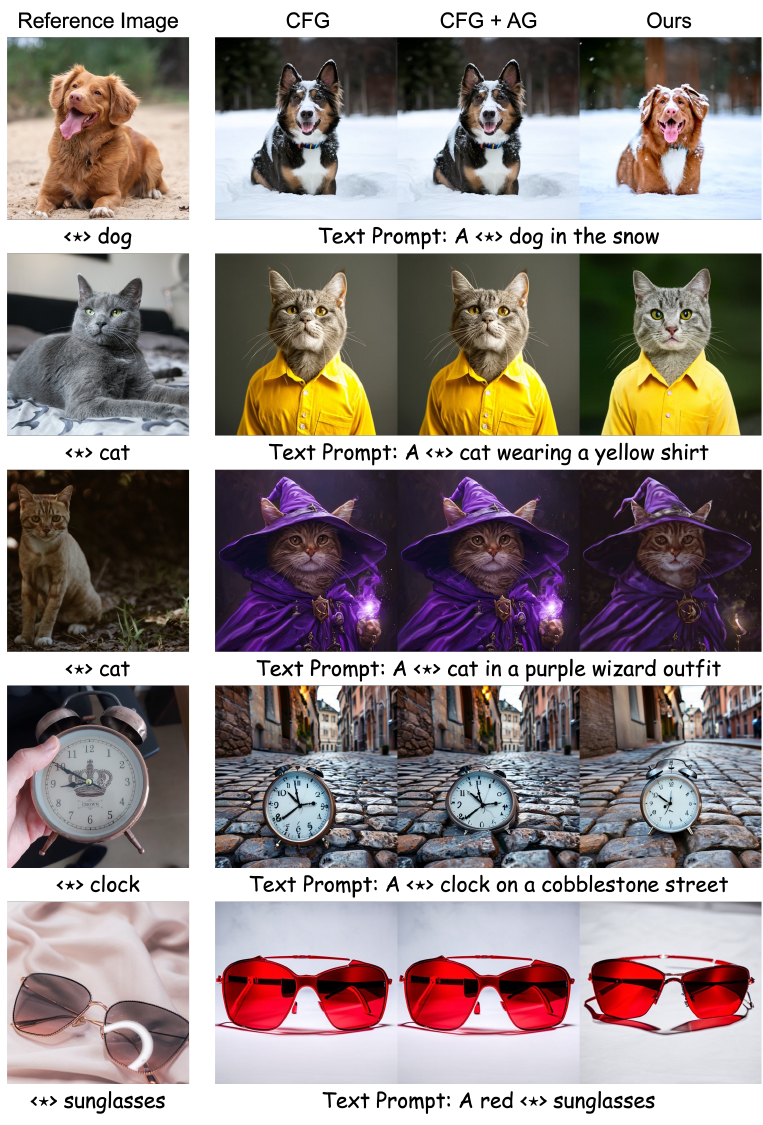}
    % \vspace{-0.5cm}
    \caption{Comparison with other guidance techniques. These images are generated by fine-tuned SANA~\cite{sana} using DB-LoRA.}
    % \vspace{-0.2cm}
    \label{supp-fig:comparison_sana}
\end{figure*}

\noindent\textbf{Additional Qualitative Results.}
Fig.~\ref{supp-fig:comparison_classdiffusion} and Fig.~\ref{supp-fig:comparison_sana} present additional qualitative results, which demonstrate our guidance's capability of improving subject and text fidelity.
These results are generated using the same weights and the same noise seed.

\section{Discussion}

\noindent\textbf{Dependency on Fine-tuned Models.}
Our approach relies on the pre-trained model as a weak model to guide the generation toward the direction of the fine-tuned model.
As a result, the effectiveness of our method is inherently tied to the quality and learning direction of the fine-tuned model.
Specifically, our guidance is most effective when the fine-tuned model has learned a more desirable distribution than the pre-trained model.
This dependency highlights the importance of optimizing fine-tuning strategies to ensure robust and meaningful personalization.

\begin{table}[t!]
\centering
\setlength{\tabcolsep}{4.5pt} % Reduce column padding
\small
\begin{tabular}{c|c|ccc}
    \toprule
    \textbf{Base Model} & \textbf{$\omega$} & \textbf{DINO} & \textbf{CLIP-I} & \textbf{CLIP-T} \\
    \midrule
    \multirow{10}{*}{\makecell{\textbf{SD 1.5} \\ \textbf{(DB-LoRA)}}}
    & 0.0 & 0.4985 & 0.7516 & 0.3260 \\
    & 0.1 & 0.4987 & 0.7514 & 0.3261 \\
    & 0.2 & 0.4991 & 0.7516 & 0.3264 \\
    & 0.3 & 0.5000 & \textbf{0.7519} & 0.3265 \\
    & 0.4 & 0.5011 & 0.7514 & 0.3271 \\
    & 0.5 & 0.5021 & 0.7512 & 0.3280 \\
    & 0.6 & \textbf{0.5024} & 0.7509 & 0.3288 \\
    & 0.7 & 0.5017 & 0.7495 & 0.3302 \\
    & 0.8 & 0.4979 & 0.7473 & 0.3320 \\
    & 0.9 & 0.4881 & 0.7425 & 0.3333 \\
    & 1.0 & 0.4701 & 0.7349 & \textbf{0.3345} \\
    \bottomrule
\end{tabular}
\vspace{-0.2cm}
\caption{Ablation study on $\omega$ using SD 1.5 with DB-LoRA.}
\label{table:omega_ablation_sd15}
\end{table}

\begin{table}[t!]
\centering
\setlength{\tabcolsep}{4.5pt} % Reduce column padding
\small
\begin{tabular}{c|c|ccc}
    \toprule
    \textbf{Base Model} & \textbf{$\omega$} & \textbf{DINO} & \textbf{CLIP-I} & \textbf{CLIP-T} \\
    \midrule
    \multirow{10}{*}{\makecell{\textbf{SD 2.1} \\ \textbf{(DB-LoRA)}}}
    & 0.0 & 0.5248 & 0.7655 & 0.3254 \\
    & 0.1 & 0.5249 & 0.7655 & 0.3253 \\
    & 0.2 & 0.5253 & 0.7656 & 0.3254 \\
    & 0.3 & 0.5260 & 0.7656 & 0.3259 \\
    & 0.4 & 0.5268 & 0.7658 & 0.3263 \\
    & 0.5 & 0.5275 & 0.7658 & 0.3270 \\
    & 0.6 & \textbf{0.5281} & \textbf{0.7660} & 0.3277 \\
    & 0.7 & 0.5272 & 0.7648 & 0.3288 \\
    & 0.8 & 0.5240 & 0.7626 & 0.3299 \\
    & 0.9 & 0.5160 & 0.7591 & 0.3311 \\
    & 1.0 & 0.4976 & 0.7519 & \textbf{0.3323} \\
    \bottomrule
\end{tabular}
\vspace{-0.2cm}
\caption{Ablation study on $\omega$ using SD 2.1 with DB-LoRA.}
\label{table:omega_ablation_sd21}
\end{table}

\begin{table}[t!]
\centering
\setlength{\tabcolsep}{4.5pt} % Reduce column padding
\small
\begin{tabular}{c|c|ccc}
    \toprule
    \textbf{Base Model} & \textbf{$\omega$} & \textbf{DINO} & \textbf{CLIP-I} & \textbf{CLIP-T} \\
    \midrule
    \multirow{10}{*}{\makecell{\textbf{SANA} \\ \textbf{(DB-LoRA)}}}
    & 0.0 & \textbf{0.5366} & \textbf{0.7522} & 0.3357 \\
    & 0.1 & 0.5341 & 0.7512 & 0.3359 \\
    & 0.2 & 0.5308 & 0.7498 & 0.3362 \\
    & 0.3 & 0.5270 & 0.7485 & 0.3364 \\
    & 0.4 & 0.5225 & 0.7474 & 0.3365 \\
    & 0.5 & 0.5178 & 0.7455 & 0.3365 \\
    & 0.6 & 0.5120 & 0.7439 & 0.3367 \\
    & 0.7 & 0.5057 & 0.7420 & \textbf{0.3368} \\
    & 0.8 & 0.4986 & 0.7396 & \textbf{0.3368} \\
    & 0.9 & 0.4906 & 0.7369 & 0.3367 \\
    & 1.0 & 0.4819 & 0.7291 & 0.3363 \\
    \bottomrule
\end{tabular}
\vspace{-0.2cm}
\caption{Ablation study on $\omega$ using SANA with DB-LoRA.}
\label{table:omega_ablation_sana}
\end{table}

\begin{table}[t!]
\centering
\setlength{\tabcolsep}{4.5pt} % Reduce column padding
\small
\begin{tabular}{c|c|ccc}
    \toprule
    \textbf{Base Model} & \textbf{$\omega$} & \textbf{DINO} & \textbf{CLIP-I} & \textbf{CLIP-T} \\
    \midrule
    \multirow{11}{*}{\makecell{\textbf{SD 1.5} \\ \textbf{(DB-LoRA + TI)}}}
    & 0.0 & 0.4814 & 0.7459 & 0.3233 \\
    & 0.1 & 0.4815 & 0.7459 & 0.3236 \\
    & 0.2 & 0.4821 & 0.7457 & 0.3239 \\
    & 0.3 & 0.4832 & 0.7456 & 0.3242 \\
    & 0.4 & 0.4847 & 0.7460 & 0.3247 \\
    & 0.5 & 0.4860 & 0.7461 & 0.3254 \\
    & 0.6 & 0.4874 & 0.7454 & 0.3260 \\
    & 0.7 & \textbf{0.4880} & \textbf{0.7461} & 0.3272 \\
    & 0.8 & 0.4867 & 0.7416 & 0.3287 \\
    & 0.9 & 0.4794 & 0.7375 & 0.3306 \\
    & 1.0 & 0.4618 & 0.7292 & \textbf{0.3316} \\
    \bottomrule
\end{tabular}
\vspace{-0.2cm}
\caption{Ablation study on $\omega$ using SD 1.5 with DB-LoRA + TI.}
\label{table:omega_ablation_sd15_ti}
\end{table}

\begin{table}[t!]
\centering
\setlength{\tabcolsep}{4.5pt} % Reduce column padding
\small
\begin{tabular}{c|c|ccc}
    \toprule
    \textbf{Base Model} & \textbf{$\omega$} & \textbf{DINO} & \textbf{CLIP-I} & \textbf{CLIP-T} \\
    \midrule
    \multirow{11}{*}{\makecell{\textbf{SD 2.1} \\ \textbf{(DB-LoRA + TI)}}}
    & 0.0 & 0.5683 & 0.7918 & 0.3163 \\
    & 0.1 & 0.5683 & 0.7917 & 0.3164 \\
    & 0.2 & 0.5684 & 0.7913 & 0.3164 \\
    & 0.3 & 0.5688 & 0.7915 & 0.3167 \\
    & 0.4 & 0.5688 & 0.7916 & 0.3172 \\
    & 0.5 & \textbf{0.5689} & 0.7913 & 0.3177 \\
    & 0.6 & \textbf{0.5689} & \textbf{0.7919} & 0.3185 \\
    & 0.7 & 0.5669 & 0.7902 & 0.3194 \\
    & 0.8 & 0.5622 & 0.7877 & 0.3207 \\
    & 0.9 & 0.5523 & 0.7832 & 0.3229 \\
    & 1.0 & 0.5291 & 0.7749 & \textbf{0.3244} \\
    \bottomrule
\end{tabular}
\vspace{-0.2cm}
\caption{Ablation study on $\omega$ using SD 2.1 with DB-LoRA + TI.}
\label{table:omega_ablation_sd21_ti}
\end{table}

\begin{table*}[t!]
\centering
\small
\begin{tabular}{c|c}
    \toprule
    \textbf{LIVE Prompt List} & \textbf{Non-LIVE Prompt List} \\
    \midrule
    'a <$\ast$> <subject> in the jungle' & 'a <$\ast$> <subject> in the jungle' \\
    'a <$\ast$> <subject> in the snow' & 'a <$\ast$> <subject> in the snow' \\
    'a <$\ast$> <subject> on the beach' & 'a <$\ast$> <subject> on the beach' \\
    'a <$\ast$> <subject> on a cobblestone street' & 'a <$\ast$> <subject> on a cobblestone street' \\
    'a <$\ast$> <subject> on top of pink fabric' & 'a <$\ast$> <subject> on top of pink fabric' \\
    'a <$\ast$> <subject> on top of a wooden floor' & 'a <$\ast$> <subject> on top of a wooden floor' \\
    'a <$\ast$> <subject> with a city in the background' & 'a <$\ast$> <subject> with a city in the background' \\
    'a <$\ast$> <subject> with a mountain in the background' & 'a <$\ast$> <subject> with a mountain in the background' \\
    'a <$\ast$> <subject> with a blue house in the background' & 'a <$\ast$> <subject> with a blue house in the background' \\
    'a <$\ast$> <subject> on top of a purple rug in a forest' & 'a <$\ast$> <subject> on top of a purple rug in a forest' \\
    'a <$\ast$> <subject> wearing a red hat' & 'a <$\ast$> <subject> with a wheat field in the background' \\
    'a <$\ast$> <subject> wearing a santa hat' & 'a <$\ast$> <subject> with a tree and autumn leaves in the background' \\
    'a <$\ast$> <subject> wearing a rainbow scarf' & 'a <$\ast$> <subject> with the Eiffel Tower in the background' \\
    'a <$\ast$> <subject> wearing a black top hat and a monocle' & 'a <$\ast$> <subject> floating on top of water' \\
    'a <$\ast$> <subject> in a chef outfit' & 'a <$\ast$> <subject> floating in an ocean of milk' \\
    'a <$\ast$> <subject> in a firefighter outfit' & 'a <$\ast$> <subject> on top of green grass with sunflowers around it' \\
    'a <$\ast$> <subject> in a police outfit' & 'a <$\ast$> <subject> on top of a mirror' \\
    'a <$\ast$> <subject> wearing pink glasses' & 'a <$\ast$> <subject> on top of the sidewalk in a crowded street' \\
    'a <$\ast$> <subject> wearing a yellow shirt' & 'a <$\ast$> <subject> on top of a dirt road' \\
    'a <$\ast$> <subject> in a purple wizard outfit' & 'a <$\ast$> <subject> on top of a white rug' \\
    'a red <$\ast$> <subject>' & 'a red <$\ast$> <subject>' \\
    'a purple <$\ast$> <subject>' & 'a purple <$\ast$> <subject>' \\
    'a shiny <$\ast$> <subject>' & 'a shiny <$\ast$> <subject>' \\
    'a wet <$\ast$> <subject>' & 'a wet <$\ast$> <subject>' \\
    'a <$\ast$> <subject> with Japanese modern city street in the background' & 'a <$\ast$> <subject> with Japanese modern city street in the background' \\
    'a <$\ast$> <subject> with a landscape from the Moon' & 'a <$\ast$> <subject> with a landscape from the Moon' \\
    'a <$\ast$> <subject> among the skyscrapers in New York city' & 'a <$\ast$> <subject> among the skyscrapers in New York city' \\
    'a <$\ast$> <subject> with a beautiful sunset' & 'a <$\ast$> <subject> with a beautiful sunset' \\
    'a <$\ast$> <subject> in a movie theater' & 'a <$\ast$> <subject> in a movie theater' \\
    'a <$\ast$> <subject> in a luxurious interior living room' & 'a <$\ast$> <subject> in a luxurious interior living room' \\
    'a <$\ast$> <subject> in a dream of a distant galaxy' & 'a <$\ast$> <subject> in a dream of a distant galaxy' \\
    \bottomrule
\end{tabular}
\vspace{-0.2cm}
\caption{Evaluation prompt list we used.}
\label{table:prompt_list}
\end{table*}

\clearpage

% \omega의 변화에 따른 다양한 모델들의 실제 수치 값들을 표로 정리해놓은 것들
% DB-LoRA 3가지 / ClassDiffusion / DB-LoRA+TI

% 정성결과 추가 (SD1.5, SD2.1, SANA)
% - User Study 때 썼던 결과 우선적으로 넣기
% - SD 1.5는 없긴 한데... 어떻게 해야할지

% 만약 DB-LoRA 같은걸로 학습할때 다른 정보가 더 끼어드는경우...?
% 

\end{document}